\documentclass[11pt]{article}

\usepackage[preprint]{acl}
\usepackage{amsmath}
\usepackage{amssymb}
\usepackage{multirow} 
\usepackage{times}
\usepackage{latexsym}
\usepackage{booktabs}
\usepackage[T1]{fontenc}
\usepackage{xcolor}
\usepackage[utf8]{inputenc}

\usepackage{microtype}

\usepackage{inconsolata}

\usepackage{graphicx}

\title{External Observers May See More Clearly: Cross-Model Span-Level Hallucination Detection in Large Language Models via Hidden State Probing}

\author{
  \textbf{Kingshuk Gupta}\textsuperscript{1,3}, \textbf{Davide Buscaldi}\textsuperscript{1,2} \\
  \textsuperscript{1}LIX, École Polytechnique \\
  \textsuperscript{2}LIPN, Sorbonne Paris Nord \\
  \textsuperscript{3}School of Electrical and Electronic Engineering, Nanyang Technological University \\  
  \\
  \small{
    \textbf{Correspondence:} \href{mailto:KINGSHUK002@e.ntu.edu.sg}{KINGSHUK002@e.ntu.edu.sg}
  }
  \texttt{KINGSHUK002@e.ntu.edu.sg}
}

\begin{document}
\maketitle
\begin{abstract}
As Large Language Models (LLMs) increasingly serve as foundational reasoning engines, their tendency to hallucinate remains a critical vulnerability. While recent internal state probes offer a promising alternative to slow external retrieval systems, they largely reduce hallucination detection to a token-wise binary classification task, failing to capture the structured, sequential boundaries of semantic drift. Here, we introduce an internal hidden state framework for fine-grained, span-level hallucination detection. By inspecting layer-wise activation patterns, we attempt to detect the exact hallucination onset and continuation tokens in an LLM generation. Our experiments show that this approach successfully isolates hallucination onsets, achieving substantial improvements in Precision-Recall AUC over random baselines despite extreme class imbalance. Ultimately, we propose a novel cross-model detection framework in which one model observes the internal representations elicited by another model's generation. We find that an external observer can match or exceed a generator's self-detection of its own hallucination onsets, including when the observer is the smaller model, suggesting that self-detection is not the ceiling for onset localisation.
\end{abstract}

\section{Introduction}

In recent years, Large Language Models (LLMs) have achieved unprecedented performance across natural language tasks, yet their persistent tendency to hallucinate limits their deployment in critical domains such as healthcare and law. To mitigate this, state of the art approaches primarily rely on external knowledge retrieval, such as RAG \citep{min2023factscore, wu2023ragtruth}. While highly effective for cloud-based models with unrestricted internet access, these external dependencies introduce severe latency and render hallucination mitigation practically impossible for local, privacy preserving LLM deployments where online search is unavailable.

Consequently, recent research has shifted toward internal state analysis, demonstrating that valid truth signals are encoded directly within a model's hidden states during inference \citep{levinstein2024simple, li2024mind}. However, existing frameworks are constrained by their architectural assumptions. Most prominent studies rely on independent probing techniques (e.g., simple MLPs or linear regression) that reduce hallucination detection to isolated, token-wise binary classification. This fundamentally fails to capture the structured, sequential dependencies of semantic drift. We argue that practical, real-time mitigation requires \textit{span-level analysis} to identify the precise temporal boundaries of a hallucination.

To achieve this, we adopt a targeted self and cross-model white-box approach. Building upon the ``Ghost in the Transformer'' framework \citep{ghostmain2025}, which established that abnormal model behaviors manifest as distinct geometric deviations, we extract high-dimensional forensic features exclusively from the model's critical layers. Our findings suggest that this ``Ghost'' signal is not merely a sentence level binary flag, but a continuous semantic signal. By projecting these internal activations into progressive sequence-labeling architectures, we utilize these geometric deviations to structurally pinpoint the exact onset of semantic drift.

\noindent Our core contributions are as follows:
\begin{itemize}
    \item \textbf{From Binary to Span-Level Detection:} We extend hidden state forensics from sentence level classification to fine-grained, token-level sequence labeling. By projecting internal states through temporally-aware architectures, we demonstrate that LLMs contain sufficient granular information to track the precise boundaries of a hallucination in real-time.\footnote{Code and datasets will be made publicly available upon publication.}

    \item \textbf{Cross-Model Supervision:} We introduce a novel cross model detection framework where one model detects hallucination spans in another model's generation from its internal representations. We show that a capable external observer can exceed self-detection of hallucination onsets, and that this holds even when the observer is smaller than the generator.
\end{itemize}
\section{Related Work}

\paragraph{Ghost in Transformer approach}: Our approach builds upon the 'Hidden State Forensics' framework \citep{ghostmain2025}, which demonstrated that abnormal LLM behaviors manifest as geometric deviations within hidden layers. While they framed hallucination detection as a system-level binary classification task, we extend this paradigm to the span-level. Instead of merely flagging an output as anomalous, we leverage these continuous forensic signals for sequence labeling to pinpoint the exact onset and termination of a hallucination.

\paragraph{Hallucination Span detection using Attention Features} : Closely related to our methodology, \citet{ogasa2025span} recently explored multi-view attention features to perform token-level hallucination detection. While employing similar classifier architectures, such as a Transformer encoder with a CRF prediction head, their framework ultimately again reduces to a token-wise binary classification task.  This formulation falls short of our attempt to precisely pinpointing the exact onset and continuous boundary of a hallucination span. Moreover, they only use internal representations from the same model while we explore cross model detection frameworks.

\paragraph{Internal-state and activation-based detectors}: Several recent methods detect hallucinations from a model's internal computation at the response level: the ICR Probe \citep{zhang2025icr} tracks the cross-layer evolution of hidden-state updates rather than static representations while semantic entropy probes \citep{kossen2024semantic} approximate semantic entropy from the hidden states of a single generation. Moving beyond probes at fixed layer token positions, ACT-ViT \citep{barshalom2025actvit} learns over full activation tensors (layers $\times$ tokens) pooled from multiple LLMs. All of these approaches target response-level judgements rather than the token-level onset and continuation boundaries that are our focus.

\paragraph{SelfCheckGPT}: \citep{manakul2023selfcheckgpt} approaches hallucination detection in a zero-resource, black-box setting by sampling multiple stochastic responses from a model and measuring the informational consistency across them. It operates on the premise that factually grounded statements are highly reproducible, whereas hallucinated facts yield contradictory variations across samples, allowing for sentence-level hallucination scoring without external databases. However, since it is sentence-level, it cannot detect sub-sentence hallucination spans.

\section{Problem Formulation}

We formulate hallucination span detection as a structured sequence labeling task, specifically predicting a ternary label for each token in the LLM-generated text to indicate whether it marks the beginning, inside, or outside of a hallucination.

Let $X = (x_1, x_2, \dots, x_n)$ represent the input prompt provided to the LLM, and $Y = (y_1, y_2, \dots, y_t)$ represent the generated response of length $t$. Our primary objective is to map the sequence of generated tokens $Y$ to a corresponding sequence of discrete span labels $Z = (z_1, z_2, \dots, z_t)$ based on the hidden states of these tokens in the LLM.

To accurately capture the boundaries of the hallucination, we define the label space using the BIO (Begin, Inside, Outside) tagging scheme. We aim to train an observer model $f_{\theta}$ that acts as a mapping function to predict a discrete label $z_t = f_\theta(y_t)$ for each token step, where $z_t \in \{0, 1, 2\}$. In this mapping, 0 represents factual (Outside HALL \footnote{HALL in this space refers to Hallucination}), 1 indicates the precise onset token of a hallucinated span (B-HALL), and 2 denotes the continuation of the hallucinated span (I-HALL). This formulation allows us to precisely identify where a hallucination starts and ends in real-time.

\section{Methodology}

\begin{figure*}[t]
    \centering
    \includegraphics[width=0.9\textwidth]{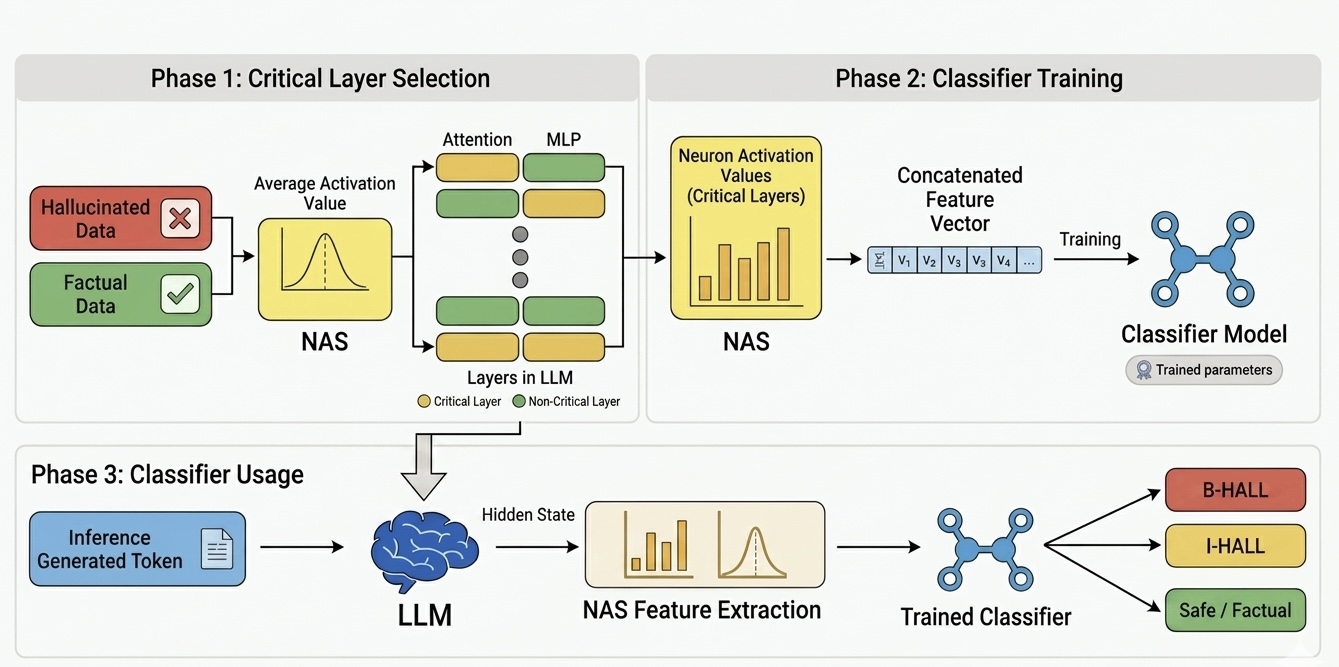}
    \caption{Token-Level Hallucination Span Detection Framework. Phase 1 identifies critical layers offline to supply the extraction targets for Phase 2, while Phase 2 processes this continuous feature sequence through the trained classifier to output structurally valid span tags}
    \label{fig:methodology_pipeline}
\end{figure*}
Modern Large Language Models distribute semantic processing across dozens of stacked transformer blocks, iteratively refining the representation of each generated token to compute a final probability distribution. Inspired by the findings of \citet{ghostmain2025}, our methodology adopts a targeted, white-box approach. Rather than analyzing the hidden states at each layer of the entire network, we first focus on identifying the specific ``critical layers'' where these factual deviations originate. Following this, we extract highly discriminative feature vectors based on localized neuron activations, which are subsequently projected into our custom sequence labeling architectures to pinpoint the precise boundaries of a hallucination.

\subsection{Critical Layer Selection and Analysis}

To isolate the specific regions of the network most sensitive to semantic drift, we systematically evaluate the activation space of the model. We sample intermediate hidden states from both factual and hallucinated text generations across a balanced calibration set. For a given token $x$, we extract the hidden states $H^{Attn}_i(x)$ and $H^{MLP}_i(x)$ at transformer block $i$.

We then aggregate these states across our respective datasets to form representative feature vectors. Formally, for a dataset $\mathcal{D}$ containing either factual (Normal) or hallucinated (Abnormal) tokens, the aggregated feature vector $F_i$ for a given layer is computed as the mean of the extracted hidden states:
\begin{equation}
    F_i = \frac{1}{|\mathcal{D}|} \sum_{x \in \mathcal{D}} H_i(x)
\end{equation}

Once we establish these representative vectors for both behaviors, we quantify the divergence between them. To do this, we compute the cosine similarity between the normal ($F^{N}_i$) and abnormal ($F^{A}_i$) feature vectors at each layer:
\begin{equation}
\begin{split}
    S_i &= \cos(F^{N}_i, F^{A}_i) 
\end{split}
\end{equation}

Because cosine similarity measures the angle between vectors in high-dimensional space, a lower value indicates a massive geometric deviation in how the layer processes factual versus hallucinated context. We calculate these similarity scores separately for all Attention and MLP modules across the $L$ blocks of the transformer, yielding two sets of scores: $\{S^{Attn}_i\}_{i=1}^L$ and $\{S^{MLP}_i\}_{i=1}^L$.

To formally rank the layers based on their discriminative power, we sort these scores in ascending order:
\begin{equation}
\begin{aligned}
    RANK_{Attn} &= \text{argsort}(\{S^{Attn}_i\}_{i=1}^L) \\
    RANK_{MLP} &= \text{argsort}(\{S^{MLP}_i\}_{i=1}^L)
\end{aligned}
\end{equation}

From these ranked lists, we select the top $\alpha$ proportion of Attention layers and the top $\beta$ proportion of MLP layers (where $\alpha, \beta \in [0, 1]$). The union of these selected layers forms our critical layer set, $\mathcal{C}_{critical}$, ensuring our downstream sequence labeler focuses exclusively on the most informative forensic signals\footnote{We revisit this design choice in Appendix~\ref{sec:layer_robustness} in light of a post-hoc robustness check.}:
\begin{equation}
\begin{split}
    \mathcal{C}_{critical} =& \ \{l^{Attn}_1, \dots, l^{Attn}_{\alpha \cdot L}\} \\ 
    &\cup \{l^{MLP}_1, \dots, l^{MLP}_{\beta \cdot L}\}
\end{split}
\end{equation}

By discarding the layers that process general syntax and focusing entirely on $\mathcal{C}_{critical}$, we significantly reduce the computational overhead for our subsequent token-level feature extraction.


\subsection{Feature Extraction using Neuron Activation Score (NAS)}

Once the optimal critical layers are isolated, we extract their raw hidden states to construct structured feature vectors for our downstream sequence-labeling architectures. To capture the intensity of the semantic drift, we utilize the \textbf{Neuron Activation Score (NAS) } as our primary feature representation.
\begin{figure*}[t]
    \centering
    \includegraphics[width=\textwidth]{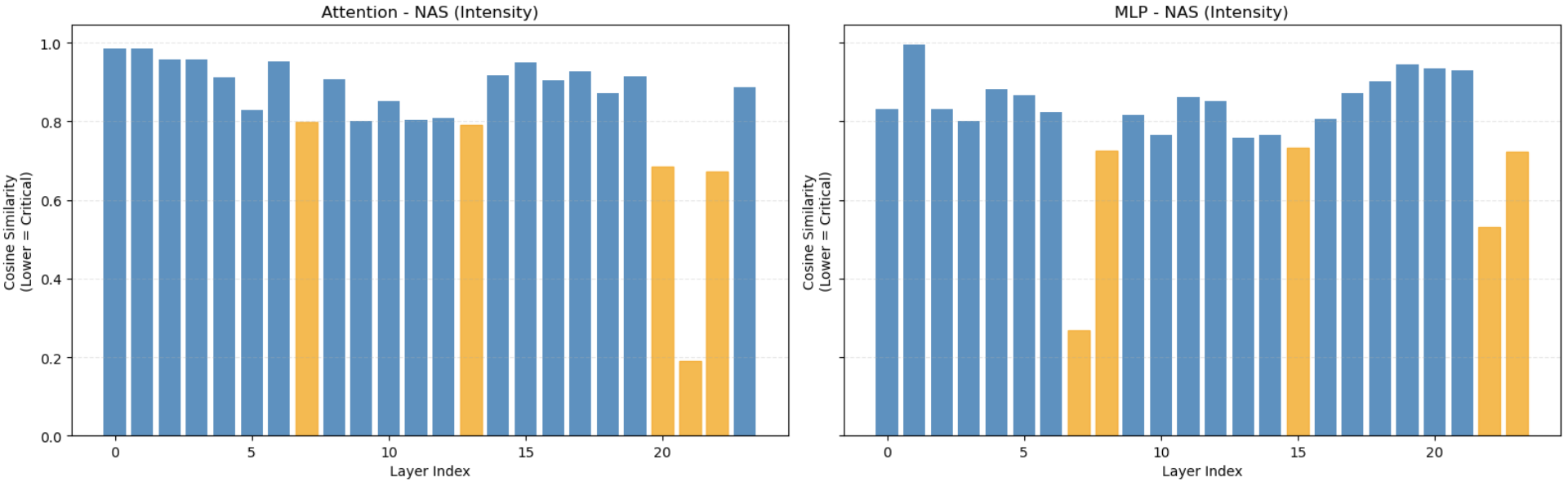}
    \caption{Critical layer analysis of the SmolLM2 (1.7B) architecture across its Attention and MLP blocks. The bars represent the cosine similarity between normal and hallucinated feature vectors at each layer depth. Layers highlighted in yellow denote the selected critical layers ($\mathcal{C}_{critical}$), where sharp geometric deviations reveal the onset of semantic drift.}
    \label{fig:critical_layer_smollm2}
\end{figure*}
The Neuron Activation Score (NAS) captures the raw intensity of each neuron’s activation, providing a direct numerical reflection of how the model is reacting to the current generation context. We leverage this feature because when an LLM drifts into a hallucinated state, it engages atypical processing pathways, causing specific neurons within the deeper layers to exhibit abnormal activation spikes or drops. By encapsulating these internal state fluctuations, the NAS serves as a precise, numerical measure of semantic deviation.

Formally, for a given generated token $x$, we define the NAS directly as the raw activation vector of the hidden state $H_i(x) \in \mathbb{R}^{d_i}$ at a critical block $i$, where $d_i$ represents the dimensionality of that hidden layer. 

To construct the final representation for the token, we extract the NAS from every module identified in our critical set $\mathcal{C}_{critical}$. By concatenating these individual layer vectors, we form a single, comprehensive, high-dimensional feature vector for that specific time step:
\begin{equation}
    F_{act}(x) = \text{Concat} \left( [H_i(x)]_{i \in \mathcal{C}_{critical}} \right)
\end{equation}

By extracting this $F_{act}(x)$ vector continuously for every single generated token, we yield a rich, temporal sequence of internal activation states. It is this sequential stream of concatenated NAS vectors that we feed into our classifier models to track hallucination spans in real-time.

\subsection{Classifier Design and Training}

To map the extracted forensic feature vectors $F_{act}(x)$ into our discrete BIO tagging space (\texttt{Safe}, \texttt{B-HALL}, \texttt{I-HALL}), we design and train a progressive series of classifier architectures, evaluating how effectively different modeling paradigms capture semantic drift.

\subsubsection{MLP Probe and Projector}

As an initial baseline, we implement a context-blind, token-wise 5-layer MLP probe that evaluates the feature vector $F_{act}(x_{t})$ for each token independently. Although the MLP lacks temporal memory, it serves as a robust baseline because the LLM's own self-attention has already aggregated the preceding sequence history into each token's hidden state. The truncated MLP additionally serves as a frozen projector, compressing the raw feature vectors for efficient ingestion by our subsequent temporal architectures.

\subsubsection{Temporal Modeling (BiLSTM-CRF)}

To explicitly track semantic drift across time, we process the projected outputs through a Bidirectional LSTM (BiLSTM), contextualizing each token on both preceding factual grounding and subsequent grammatical flow. However, predicting labels independently from the BiLSTM's hidden states can yield structurally invalid sequences, such as an \texttt{I-HALL} continuation without a preceding \texttt{B-HALL} onset. We therefore append a Conditional Random Field (CRF) layer, which models transition probabilities globally and applies Viterbi decoding to enforce a logically consistent tag sequence.

\subsubsection{Adding Attention Mechanism (BiLSTM-Attn-CRF)}

Recurrent architectures suffer from information decay over long sequences, causing them to miss long-range semantic contradictions. To overcome this, we embed a Multi-Head Self-Attention layer after the BiLSTM, allowing the network to cross-reference the current token against the entire generated sequence. A residual connection across the attention block preserves the syntactic features learned by the recurrent layer, and the enriched representation is then decoded by the final CRF head.

\subsubsection{BERT-CRF}

To fully leverage the forensic signals without the information loss of aggressive compression, our final architecture pairs a custom LayerNorm projector with a pre-trained BERT encoder \citep{devlin2018bert} and a CRF head. The projector bridges the domain gap by aligning the continuous LLM hidden states into BERT's native embedding space. Although BERT is pre-trained on natural language, its deep bidirectional attention effectively models the sequential dependencies in these activation streams, and the contextualized outputs are decoded by the CRF layer into the final span tags.
\section{Evaluation}

\subsection{Datasets}
To evaluate hallucination dynamics across varying parameter scales, we extract span-level annotated hidden states from three distinct white-box setups: a custom PsiloQA \cite{rykov2025modelslielearnmultilingual} generation using SmolLM2 (1.7B) \citep{allal2025smollm2}, the original PsiloQA dataset with TinyLlama (1.1B)\citep{liao2024tinyllama}, and a specialized RAGTruth QA \citep{wu2023ragtruth} subset using Mistral-7B-Instruct \citep{jiang2023mistral}. The token-level distribution for these combined corpora is detailed in Table \ref{tab:dataset_distribution}.

\begin{table}[ht]
\centering
\small
\setlength{\tabcolsep}{2.2pt}
\begin{tabular}{@{}lcccc@{}}
\toprule
\textbf{Dataset (Model)} & \textbf{Safe (O)} & \textbf{B-HALL} & \textbf{I-HALL} & \textbf{Total} \\ \midrule
PsiloQA (SmolLM2)         & 7,797             & 265             & 2,923           & 10,985         \\
PsiloQA (TinyLlama)       & 13,022            & 707             & 18,288          & 32,017         \\
RAGTruth (Mistral)       & 11,241            & 57              & 616             & 11,914         \\ \bottomrule
\end{tabular}
\caption{Token-level distribution of the BIO tags across the test sets of the three evaluated datasets. Each dataset was uniformly partitioned into standard 80/20 train/test splits prior to hidden state extraction}
\label{tab:dataset_distribution}
\end{table}

\begin{table*}[t]
\centering
\small
\setlength{\tabcolsep}{5pt}
\begin{tabular}{@{}l|ccc|ccc|ccc@{}}
\toprule
& \multicolumn{3}{c|}{\textbf{SmolLM2 (1.7B)}} & \multicolumn{3}{c|}{\textbf{TinyLlama (1.1B)}} & \multicolumn{3}{c}{\textbf{Mistral (7B)}} \\
\textbf{Architecture} & \textbf{P} & \textbf{R} & \textbf{F1} & \textbf{P} & \textbf{R} & \textbf{F1} & \textbf{P} & \textbf{R} & \textbf{F1} \\ \midrule
\multicolumn{10}{l}{\textit{\textbf{Onset Token (\texttt{B-HALL})}}} \\ \midrule
MLP Baseline & 0.32 & 0.39 & \textbf{0.35} & 0.29 & 0.36 & 0.32 & 0.19 & \textbf{0.26} & 0.22 \\
BiLSTM & 0.29 & \textbf{0.40} & 0.33 & 0.28 & \textbf{0.38} & 0.32 & 0.25 & 0.25 & 0.25 \\
BiLSTM-CRF & 0.42 & 0.24 & 0.31 & 0.32 & 0.30 & 0.31 & 0.38 & 0.18 & 0.24 \\
BiLSTM-Attn-CRF & 0.42 & 0.30 & \textbf{0.35} & \textbf{0.41} & 0.28 & \textbf{0.34} & \textbf{0.41} & 0.21 & \textbf{0.28} \\
BERT-CRF (Ours) & \textbf{0.54} & 0.23 & 0.33 & \textbf{0.41} & 0.30 & \textbf{0.34} & 0.25 & 0.07 & 0.11 \\ \midrule
\multicolumn{10}{l}{\textit{\textbf{Continuation Tokens (\texttt{I-HALL})}}} \\ \midrule
MLP Baseline & 0.49 & 0.49 & 0.49 & 0.71 & \textbf{0.90} & \textbf{0.80} & 0.49 & \textbf{0.42} & \textbf{0.46} \\
BiLSTM & 0.47 & \textbf{0.53} & \textbf{0.50} & 0.74 & 0.84 & 0.79 & 0.69 & 0.30 & 0.42 \\
BiLSTM-CRF & \textbf{0.52} & 0.35 & 0.42 & \textbf{0.75} & 0.83 & 0.79 & \textbf{0.70} & 0.27 & 0.39 \\
BiLSTM-Attn-CRF & \textbf{0.52} & 0.40 & 0.45 & \textbf{0.75} & 0.84 & 0.79 & 0.69 & 0.28 & 0.40 \\
BERT-CRF (Ours) & 0.50 & 0.48 & 0.49 & \textbf{0.75} & 0.85 & \textbf{0.80} & 0.52 & 0.11 & 0.18 \\ \bottomrule
\end{tabular}
\caption{Intra-model span detection performance across three LLM architectures. Precision (P), Recall (R), and F1-scores are reported exclusively for the minority hallucination classes to highlight the detection of semantic drift. The \texttt{Safe (O)} majority class is excluded to prevent metric inflation. All experiments were conducted using the top-5 Attention and top-5 MLP critical layers.}
\label{tab:main_results}
\end{table*}
\subsection{Baselines}
Given our formulation of hallucination detection as a token-level sequence labeling task, sentence and document level hidden-state probes such as \citet{levinstein2024simple} and the response-level internal-state detectors discussed in Section 2, extract a single factuality score per claim, and cannot be mapped onto our token-level sequence space.

Our evaluation therefore proceeds along two fronts. First, we establish internal architectural ablations, comparing a context-blind \textbf{5-layer MLP probe} and a \textbf{standard BiLSTM} against progressively more complex architectures, to identify which underlying mechanism best captures the sequential progression of semantic drift. Second, to situate our framework against existing span-level work, we compare against the multi-view attention baseline of \citet{ogasa2025span}, which feeds Average Incoming Attention together with attention-entropy views into a Transformer encoder with a CRF prediction head. We reimplement this baseline and adapt it to our ternary BIO onset task, evaluating it under the same label alignment and imbalance handling as our own framework.

\subsection{Experimental Results}

Table \ref{tab:main_results} presents the intra-model span detection performance across the three evaluated LLMs. A primary global trend emerges regarding the context-blind MLP baseline: across all generators, it achieves surprisingly competitive F1-scores for continuous hallucinated spans (e.g., reaching 0.80 for \texttt{I-HALL} on TinyLlama). This indicates that the extracted hidden states already encode rich sequential history from the LLMs' native self-attention layers. However, without structural constraints, both the MLP and the standard BiLSTM suffer from poor onset (\texttt{B-HALL}) Precision, as they frequently predict grammatically invalid tag sequences (such as an \texttt{I-HALL} continuation without a preceding \texttt{B-HALL} onset).

Integrating a Conditional Random Field (CRF) directly resolves this grammatical weakness globally. By penalizing impossible state transitions, the CRF forces the model to explicitly identify the onset boundary before tracking the span. While enforcing strict sequence validity naturally depresses raw recall, it triggers a consistent, cross-model surge in \texttt{B-HALL} Precision. For instance, transitioning from BiLSTM to BiLSTM-CRF boosts onset Precision from 0.29 to 0.42 in SmolLM2, and from 0.25 to 0.38 in Mistral.

Finally, the advanced attention-based architectures reveal distinct, model-specific ceilings. On SmolLM2, the \textbf{BERT-CRF} framework performs best, achieving the highest onset Precision (0.54). However, this same architecture plateaus on TinyLlama (tying the BiLSTM-Attn-CRF at 0.41 Precision) and struggles to extract a clear signal from Mistral's states. This divergence indicates that advanced sequence modeling does not uniformly improve onset detection across generators, and that self-detection performance varies with the generator in ways we examine in the following sections.

\subsection{Hallucinated Token Ratios and Diagnostic Metrics}

A critical factor influencing the detection performance across our evaluated models, is the extreme sparsity of hallucination onset tokens. As detailed in Table \ref{tab:dataset_distribution}, the baseline hallucination ratios vary significantly across datasets. While the \texttt{B-HALL} tokens constitute roughly 2.4\% and 2.2\% of the PsiloQA test sets for SmolLM2 and TinyLlama respectively, the RAGTruth subset for Mistral is far more imbalanced, with \texttt{B-HALL} tokens comprising only \textbf{0.48\%} (57 out of 11,914 tokens) of the generated text.

This severe class imbalance fundamentally impacts standard evaluation metrics like F1-score. In datasets like Mistral's, the overwhelming prevalence of factual tokens heavily biases the CRF's learned transition matrix. Consequently, the CRF’s overwhelming bias toward factual text easily overrides faint anomaly signals, naturally depressing recall for rare onsets. Conversely, the much larger proportion of \texttt{I-HALL} continuation tokens across all datasets naturally inflates their corresponding F1-scores (e.g., reaching 0.80 on TinyLlama), as the models have ample continuous data to track the semantic drift once the hallucination onset is identified.

To confirm the model detects true geometric anomalies as hallucinations rather than exploiting fixed thresholds or class imbalance, we require threshold-independent evaluation. Focusing on our primary dataset (PsiloQA(SmolLM2), where \texttt{B-HALL} frequency is 2.4\%), we compare the Area Under the Receiver Operating Characteristic (AUROC) and Precision-Recall Area Under the Curve (PR-AUC) of our BERT-CRF against the context-blind MLP baseline (Table \ref{tab:diagnostic_metrics}).

\begin{table}[t]
\centering
\small
\setlength{\tabcolsep}{4pt}
\resizebox{\columnwidth}{!}{%
\begin{tabular}{@{}llcc@{}}
\toprule
\textbf{Token Class} & \textbf{Model} & \textbf{AUROC} & \textbf{PR-AUC} \\ \midrule
\multirow{3}{*}{Onset (\texttt{B-HALL})} & MLP Baseline & \textbf{0.900} & 0.330 \\
& Ogasa \citep{ogasa2025span} & 0.931 & \textbf{0.371} \\
& \textbf{BERT-CRF} & 0.895 & 0.362 \\ \midrule
\multicolumn{4}{r}{\textit{Random Baseline: 0.024}} \\ \midrule\midrule
\multirow{2}{*}{Continuation (\texttt{I-HALL})} & MLP Baseline & 0.716 & \textbf{0.460} \\
& \textbf{BERT-CRF} & \textbf{0.735} & 0.446 \\ \midrule
\multicolumn{4}{r}{\textit{Random Baseline: 0.266}} \\ \bottomrule
\end{tabular}%
}
\caption{Threshold-independent onset diagnostics on the SmolLM2 test set, including the intra-model attention baseline of \citet{ogasa2025span}. The \textit{Random Baseline} equals the class prevalence. Single-seed.}
\label{tab:diagnostic_metrics}
\end{table}

The BERT-CRF achieves an onset AUROC of 0.895, effectively matching the MLP baseline at 0.900, confirming that the extracted activations support an accurate probability ranking even for a context-blind classifier. As AUROC can be overly optimistic under imbalance, we treat PR-AUC as the primary metric, where the BERT-CRF yields 0.362 against a random floor of just 0.024, a 15-fold improvement that indicates the rare onset signal is genuinely isolated rather than exploited through class priors. On the same task, the multi-view attention baseline of \citet{ogasa2025span} reaches a comparable intra-model PR-AUC of 0.371, closely matching our framework and confirming that both feature families recover the onset signal from a model's own states.

\begin{table*}[t]
\centering
\small
\setlength{\tabcolsep}{4pt}
\begin{tabular}{@{}l|ccc|ccc|ccc|ccc|ccc@{}}
\toprule
& \multicolumn{15}{c}{\textbf{Cross-Model Framework Evaluation}} \\ \cmidrule{2-16}
\textbf{Architecture} & \multicolumn{3}{c|}{$\mathbf{Qwen_{Smol}}$} & \multicolumn{3}{c|}{$\mathbf{Gemma2_{Smol}}$} & \multicolumn{3}{c|}{$\mathbf{Gemma4_{Smol}}$} & \multicolumn{3}{c|}{$\mathbf{Gemma2_{Mist}}$} & \multicolumn{3}{c}{$\mathbf{Smol_{Mist}}$} \\
& \textbf{P} & \textbf{R} & \textbf{F1} & \textbf{P} & \textbf{R} & \textbf{F1} & \textbf{P} & \textbf{R} & \textbf{F1} & \textbf{P} & \textbf{R} & \textbf{F1} & \textbf{P} & \textbf{R} & \textbf{F1} \\ \midrule
\multicolumn{16}{l}{\textit{\textbf{Onset Token (\texttt{B-HALL})}}} \\ \midrule
MLP Baseline    & 0.37 & 0.41 & 0.39 & 0.42 & \textbf{0.57} & 0.48 & 0.34 & \textbf{0.67} & 0.45 & 0.10 & 0.39 & 0.16 & 0.11 & 0.32 & 0.17 \\
BiLSTM          & 0.39 & \textbf{0.42} & \textbf{0.41} & 0.45 & 0.49 & 0.47 & 0.39 & 0.54 & 0.45 & 0.29 & \textbf{0.46} & \textbf{0.35} & 0.17 & \textbf{0.47} & 0.25 \\
BiLSTM-CRF      & \textbf{0.53} & 0.29 & 0.38 & 0.55 & 0.36 & 0.43 & \textbf{0.50} & 0.33 & 0.40 & \textbf{0.40} & 0.32 & \textbf{0.35} & 0.33 & 0.14 & 0.20 \\
BiLSTM-Attn-CRF & \textbf{0.53} & 0.29 & 0.38 & 0.57 & 0.36 & 0.44 & 0.46 & 0.26 & 0.33 & 0.38 & 0.26 & 0.31 & 0.44 & 0.19 & \textbf{0.27} \\
BERT-CRF (Ours) & \textbf{0.53} & 0.28 & 0.37 & \textbf{0.63} & 0.28 & 0.39 & 0.42 & 0.66 & \textbf{0.51} & \textbf{1.00} & 0.02 & 0.03 & \textbf{1.00} & 0.05 & 0.10 \\ \midrule
\multicolumn{16}{l}{\textit{\textbf{Continuation Tokens (\texttt{I-HALL})}}} \\ \midrule
MLP Baseline    & 0.53 & \textbf{0.47} & \textbf{0.50} & 0.48 & \textbf{0.57} & \textbf{0.52} & 0.45 & 0.56 & 0.50 & 0.43 & 0.46 & 0.45 & 0.43 & 0.31 & 0.36 \\
BiLSTM          & 0.54 & 0.41 & 0.47 & 0.51 & 0.48 & 0.49 & 0.47 & 0.47 & 0.47 & 0.63 & \textbf{0.56} & \textbf{0.59} & 0.57 & \textbf{0.38} & \textbf{0.46} \\
BiLSTM-CRF      & 0.59 & 0.30 & 0.40 & 0.52 & 0.46 & 0.49 & \textbf{0.53} & 0.35 & 0.42 & 0.60 & 0.49 & 0.54 & \textbf{0.64} & 0.32 & 0.43 \\
BiLSTM-Attn-CRF & 0.58 & 0.34 & 0.43 & \textbf{0.54} & 0.45 & 0.49 & 0.49 & 0.42 & 0.45 & \textbf{0.65} & 0.54 & \textbf{0.59} & \textbf{0.66} & 0.32 & 0.43 \\
BERT-CRF (Ours) & \textbf{0.66} & 0.28 & 0.39 & 0.46 & 0.47 & 0.47 & 0.47 & \textbf{0.59} & \textbf{0.52} & 0.64 & 0.32 & 0.42 & 0.55 & 0.34 & 0.42 \\ \bottomrule
\end{tabular}
\caption{Comprehensive cross-model hallucination span detection performance. The detection pairs $\mathbf{Qwen_{Smol}}$ utilizes Qwen (1.5B) to evaluate SmolLM2 (1.7B) generations, $\mathbf{Gemma2_{Smol}}$ and $\mathbf{Gemma4_{Smol}}$ utilize Gemma 2 (2B) and Gemma 4 (E2B) respectively to evaluate SmolLM2 (1.7B), $\mathbf{Gemma2_{Mist}}$ utilizes Gemma 2 (2B) to evaluate Mistral (7B), and $\mathbf{Smol_{Mist}}$ utilizes the smaller SmolLM2 (1.7B) to evaluate Mistral (7B). Precision (P), Recall (R), and F1-scores are reported as before.}
\label{tab:cross_model_comprehensive}
\end{table*}

\subsection{Scaling Dynamics and Probing Limits}

While our self-evaluation framework demonstrates promising results for span-level hallucination detection, performance degrades significantly when evaluating the larger Mistral (7B). This degradation is partially attributable to the extreme class imbalance in the Mistral dataset, where onset tokens constitute merely 0.48 percent of the generated text. Beyond imbalance, Mistral distributes its processing across 32 transformer blocks, and a fixed extraction budget over its critical layers captures a weaker, less concentrated onset signal than the equivalent budget yields on the smaller generators. We establish this as an empirical observation rather than a validated mechanism, as our experimental setup does not isolate depth from the architectural, scale, and pretraining differences that simultaneously distinguish these models.

The practical consequence, however, motivates the core of our work. When a generator's own hidden states yield only a faint self-detection signal, a natural question arises as to whether an external model, evaluating the same generated text, can recover the onset boundary more reliably. We investigate this cross-model setting in the following section.

\subsection{Cross-Model Transferability}

Having established that a model's own hidden states may yield only a faint self-detection signal, we investigate whether an external model, evaluating the same generated text, can recover the hallucination onset more reliably. We evaluate the cross-model transferability of our framework utilizing external supervisors like Qwen (1.5B)~\citep{bai2023qwen}, Gemma 2 (2B)~\citep{riviere2024gemma2}, and the larger Gemma 4 (E2B)~\citep{gemma2026gemma4} to evaluate generations from both SmolLM2 (1.7B) and Mistral (7B), extracting the Neuron Activation Scores (NAS) exclusively from the supervisors' critical layers.

The results, as presented in Table~\ref{tab:cross_model_comprehensive}, indicate that a capable external supervisor decodes hallucination onsets at least as effectively as the generator decodes itself. When external models act as supervisors evaluating SmolLM2's generations ($\mathbf{Qwen_{Smol}}$ and $\mathbf{Gemma2_{Smol}}$), Gemma 2 leverages our BERT-CRF framework to achieve a \texttt{B-HALL} Precision of 0.63, and Qwen decodes the signal effectively using just the intermediate BiLSTM-CRF architecture to reach a 0.53 Precision. This cross-model gain also holds against an external baseline. Using Gemma 2 (2B) to observe SmolLM2, the multi-view attention baseline of \citet{ogasa2025span} reaches an onset PR-AUC of 0.407, which our NAS framework exceeds at 0.471. 

A central concern with these comparisons is that every external observer above is more capable than the SmolLM2 generator it evaluates, confounding external observation with observer capability. To isolate this, we run the opposite configuration, tasking the smaller SmolLM2 (1.7B) with detecting the onsets of the larger Mistral (7B) generator ($\mathbf{Smol_{Mist}}$ in Table~\ref{tab:cross_model_comprehensive}), and compare it against Mistral's own self-detection under identical test folds and onset prevalence. Table~\ref{tab:self-vs-external} reports the onset PR-AUC for this comparison. Under structured CRF decoding, the smaller external observer matches or exceeds the larger model's self-detection, reaching 0.242 with the BiLSTM-Attn-CRF against Mistral's ceiling of 0.221, and 0.227 against 0.172 with the BiLSTM-CRF. Because SmolLM2 operates at lower parameter scale than Mistral, this advantage cannot be attributed to the observer merely being larger. We read this as evidence that self-detection is not the ceiling for onset localization, and that an external observer can recover onsets a model does not surface as clearly from its own states.

\begin{table}[t]
\centering
\small
\setlength{\tabcolsep}{5pt}
\begin{tabular}{lcc}
\toprule
\textbf{Architecture} & \textbf{Self} & \textbf{External} \\
 & (Mistral) & (SmolLM2$\rightarrow$Mistral) \\
\midrule
MLP             & \textbf{0.172} & 0.130 \\
BiLSTM          & \textbf{0.205} & 0.193 \\
BiLSTM-CRF      & 0.172 & \textbf{0.227} \\
BiLSTM-Attn-CRF & 0.221 & \textbf{0.242} \\
BERT-CRF        & 0.043 & \textbf{0.079} \\
\midrule
Random floor    & 0.005 & 0.005 \\
\bottomrule
\end{tabular}
\caption{Onset (\texttt{B-HALL}) PR-AUC on RAGTruth: Mistral-7B self-detection vs.\ a smaller external observer (SmolLM2-1.7B), same test fold. Bold marks the higher value per row. Single-seed.}
\label{tab:self-vs-external}
\end{table}

However, we do emphasize the limits of this evidence. The comparison rests on a single observer pair, a single dataset, and a single seed, over a test fold containing only 57 onset tokens. The external advantage further emerges specifically under CRF decoding rather than uniformly across architectures, since the context-blind MLP and the plain BiLSTM slightly favour self-detection. We therefore present this as a targeted control that removes the capability confound in one direction, rather than as a general law that external observation always exceeds self-detection.

Finally, we observe that the BERT-CRF, our heaviest architecture, is consistently the least robust under extreme onset sparsity. It records the lowest onset PR-AUC in both the self and external settings on Mistral (0.043 and 0.079), and collapses into a near-degenerate high-precision, near-zero-recall regime when Gemma 2 supervises Mistral ($\mathbf{Gemma2_{Mist}}$). This reinforces that architectural capacity does not monotonically improve onset detection, and that lighter temporal models decode the onset signal more reliably when onsets are scarce.

\section{Conclusion and Outlook}

As LLMs increasingly serve as foundational reasoning engines, mitigating their hallucinations requires moving beyond binary verification. In this work, we introduced a white-box framework for fine-grained, token-level hallucination span detection. By extracting continuous Neuron Activation Scores (NAS) from critical Transformer layers, our results suggest that semantic drift correlates with trackable shifts in layer-wise activations. Through a systematic comparison of classifier architectures, we found that enforcing sequence-level constraints improves onset precision despite extreme class imbalance.

Our cross-model framework demonstrates that a capable external observer can match and even exceed a model's self-detection of its own hallucination onsets. This holds even when the observer is smaller than the generator it evaluates, indicating that self-detection is not the ceiling for onset localization and that external observation is not merely a function of observer scale. However, the exact reasoning behind certain observers performing better at detection than the generating model itself and which architectural properties make an efficient supervisor remain open questions for future work.

\section{Limitations}

While our framework successfully maps hallucination spans, it is subject to several structural and empirical constraints. First, extracting high-dimensional hidden states across multiple transformer layers imposes an extreme burden on GPU memory, currently restricting this methodology to offline, post-hoc analysis rather than real-time edge deployment. 

Second, we recognize that our evaluation is constrained to QA-specific datasets and generative models under 7B parameters due to strict academic compute limitations. It remains an open empirical question whether our methods work across frontier models (70B+) or across long-form, open-domain generation tasks (e.g., biography generation). Consequently, we present this cross-model framework not as a fully generalized solution, but as a foundational proof-of-concept to be scaled by future work. 

A further limitation is statistical: all reported results come from a single random seed (42), and our central cross-model comparison rests on a single observer–generator pair over a test fold of only 57 onset tokens. We therefore treat these results as controlled but preliminary evidence, and leave multi-seed variance estimation and a broader range of observer–generator pairings to future work.

Finally, due to strict conference page limits, we were unable to include a comprehensive sensitivity ablation regarding the critical layer selection thresholds ($\alpha$ and $\beta$). While our standardized Top-5 layer budget proved highly effective for compact architectures, determining the optimal, dynamic extraction footprint for scaling to deeper networks remains a critical direction for subsequent research.


%

\bibliography{custom}

\appendix

\section{Reproducibility and Implementation Details}
\label{sec:appendix_reproducibility}

To ensure full transparency and reproducibility, we detail below the architecture configurations, per-dataset training hyperparameters, layer-selection thresholds, and dataset construction pipelines used throughout this study. All experiments used a fixed random seed (42).

\subsection{Classifier Architecture Configurations}
\label{sec:appendix_architectures}

All five classifiers map the concatenated NAS feature vector of dimensionality $d_{\text{in}}$, set by the generator's hidden size and the number of critical layers, into the ternary BIO tag space. Table \ref{tab:architecture_configs} summarizes the layer dimensions, which are held fixed across all datasets and observers except where noted.

\paragraph{MLP Probe and Projector}
The MLP is a five-layer feed-forward network, $d_{\text{in}} \rightarrow 1024 \rightarrow 512 \rightarrow 128 \rightarrow 64 \rightarrow 3$, with ReLU activations and dropout (0.3) after the first two hidden layers. For the temporal architectures, we truncate the final two layers of this trained network to obtain a frozen 64-dimensional projector, so that all recurrent models receive a compressed 64-dimensional input.

\paragraph{BiLSTM and BiLSTM-CRF}
The recurrent models pass the 64-dimensional projected features through a single-layer bidirectional LSTM with a hidden size of 32 per direction, yielding a 64-dimensional contextual representation that a linear layer maps to the three tag emissions. The BiLSTM-CRF appends a linear-chain CRF over these emissions and decodes with the Viterbi algorithm.

\paragraph{BiLSTM-Attn-CRF}
This variant inserts a Multi-Head Self-Attention layer (4 heads, embedding dimension 64) after the BiLSTM, with a residual connection adding the attention output back to the recurrent output before the emission and CRF layers.

\paragraph{BERT-CRF}
The projector maps the raw NAS vector through $d_{\text{in}} \rightarrow 2048 \rightarrow 768$ with a GELU activation, dropout (0.2), and LayerNorm at both input and output, producing embeddings in the native 768-dimensional space of \texttt{bert-base-uncased}. These are passed as \texttt{inputs\_embeds} to the pre-trained encoder, followed by dropout (0.3), a linear emission head, and a linear-chain CRF. For the larger TinyLlama feature dimension, the projector's intermediate width is increased to 4096.

\begin{table}[ht]
\centering
\small
\resizebox{\columnwidth}{!}{%
\begin{tabular}{@{}lll@{}}
\toprule
\textbf{Model} & \textbf{Component} & \textbf{Configuration} \\ \midrule
MLP & Hidden layers & $d_{\text{in}}$--1024--512--128--64--3 \\
 & Activation & ReLU, Dropout 0.3 \\ \midrule
BiLSTM & Input dim & 64 (projected) \\
 & LSTM hidden & 32 / direction, 1 layer \\
 & Head & Linear $64 \rightarrow 3$ \\ \midrule
BiLSTM-CRF & Base & BiLSTM (above) \\
 & Decoder & Linear-chain CRF \\ \midrule
BiLSTM-Attn-CRF & Attention & 4 heads, dim 64 \\
 & Residual & LSTM $+$ Attention \\
 & Decoder & Linear-chain CRF \\ \midrule
BERT-CRF & Projector & $d_{\text{in}}$--2048--768, GELU \\
 & Encoder & \texttt{bert-base-uncased} \\
 & Decoder & Linear-chain CRF \\ \bottomrule
\end{tabular}%
}
\caption{Layer configurations for each classifier, held fixed across datasets. $d_{\text{in}}$ is the concatenated NAS input dimension, determined by the generator's hidden size and critical-layer count. The BERT-CRF projector intermediate is widened to 4096 for TinyLlama.}
\label{tab:architecture_configs}
\end{table}

\subsection{Critical Layer Selection Robustness Check}
\label{sec:layer_robustness}

We selected critical layers using the centroid based cosine divergence heuristic of \citet{ghostmain2025}, which reported this criterion as effective for identifying anomaly-sensitive layers. Following the first review cycle, we investigated whether alternative layer choices could match or exceed it. Table~\ref{tab:layer_robustness_results} reports results from repeating the full architecture progression (MLP, BiLSTM, BiLSTM-CRF, BiLSTM-Attn-CRF) on a single, fixed random set of five attention and five MLP layers (SmolLM2, \texttt{B-HALL}).

\begin{table}[ht]
\centering
\small
\resizebox{\columnwidth}{!}{%
\begin{tabular}{@{}lccccc@{}}
\toprule
\textbf{Architecture} & \textbf{P} & \textbf{R} & \textbf{F1} & \textbf{PR-AUC} & \textbf{AUROC} \\
\midrule
MLP              & 0.360 & 0.558 & 0.438 & 0.380 & 0.854 \\
BiLSTM           & 0.323 & 0.585 & 0.416 & 0.305 & 0.800 \\
BiLSTM-CRF       & 0.492 & 0.340 & 0.402 & 0.252 & 0.784 \\
BiLSTM-Attn-CRF  & 0.517 & 0.340 & 0.410 & 0.313 & 0.783 \\
\bottomrule
\end{tabular}%
}
\caption{Onset (\texttt{B-HALL}) detection on SmolLM2 using a single fixed random layer set, not the critical layers used elsewhere in this paper. One draw, not an average over draws; single classifier seed. Corresponding critical-layer figures are given in the text below.}
\label{tab:layer_robustness_results}
\end{table}

Table~\ref{tab:layer_robustness_results} shows Precision, Recall, and F1 matching or exceeding the corresponding critical-layer figures at every architecture stage. The critical-layer figures, reported in Table~\ref{tab:main_results}, are: MLP 0.32/0.39/0.35, BiLSTM 0.29/0.40/0.33, BiLSTM-CRF 0.42/0.24/0.31, and BiLSTM-Attn-CRF 0.42/0.30/0.35 (P/R/F1). PR-AUC for the MLP probe also improved under the random layers, from 0.330 (Table~\ref{tab:diagnostic_metrics}) to 0.380, though AUROC was marginally lower (0.900 versus 0.854). PR-AUC and AUROC were not reported for the BiLSTM family in the main results, so no comparison is available for those three rows.

The precision gains from CRF decoding (Section~5.3) held under this alternative selection: adding the CRF, and then attention, raised precision in the same pattern as with the critical layers. This indicates that the framework's benefit does not depend on the specific layer-selection heuristic feeding it.

We retain the original critical-layer results as our primary reported figures, for continuity with the reviewed submission. Within the time available for this resubmission, we were not able to re-run the cross-model and cross-dataset experiments (Sections~5.5--5.6) under the alternative selection, or to average over multiple random draws rather than the single one reported here. A systematic study of layer selection, including its interaction with observer--generator pairing, is left to future work.

\subsection{Self-Evaluation: Training Strategies and Hyperparameters}

We train all classifiers with the Adam family of optimizers under strict GPU-memory constraints. The MLP is trained token-wise with a batch size of 64, while the sequence models are trained per-document to preserve span continuity. To counter the extreme onset sparsity, we apply class-weighted Cross-Entropy penalties on \texttt{B-HALL}. Because onset prevalence differs markedly across generators, these weights are tuned per dataset rather than fixed globally, with the sparsest corpora receiving the largest penalties. For CRF-based models, this weighting is applied as an auxiliary loss in parallel with the sequence CRF objective to prevent transition-matrix collapse. Representative values are given in Table \ref{tab:training_hyperparameters}.

\begin{table}[ht]
\centering
\small
\resizebox{\columnwidth}{!}{%
\begin{tabular}{@{}llp{4.2cm}@{}}
\toprule
\textbf{Hyperparameter} & \textbf{Value} & \textbf{Scope} \\ \midrule
Optimizer & Adam / AdamW & All architectures. \\
Learning rate & $1\times10^{-3}$ & Classifier heads and projectors. \\
BERT encoder LR & $2\times10^{-5}$ -- $5\times10^{-5}$ & After a 1-epoch warm-up. \\
\texttt{B-HALL} weight & 20 -- 300 & Tuned per dataset by onset sparsity. \\
\texttt{I-HALL} weight & 5 -- 30 & Tuned per dataset. \\
Training epochs & 5 -- 15 & By architecture and token volume. \\
MLP batch size & 64 & Token-wise training. \\
Sequence batch size & 1 (per doc) & Variable-length sequences. \\
ANE mask threshold & 0.2 & NAS masking. \\ \bottomrule
\end{tabular}%
}
\caption{Representative self-evaluation training hyperparameters. Class weights and epochs vary by dataset to accommodate differing onset sparsity; exact per-model values are released with our code.}
\label{tab:training_hyperparameters}
\end{table}

\paragraph{Per-Dataset Class Weighting}
The \texttt{B-HALL} penalty scales with onset sparsity. On the SmolLM2 PsiloQA data it ranges from 20 (MLP, BiLSTM) to 50 (BERT-CRF); on the sparser TinyLlama and Mistral corpora it is raised as high as 300 to prevent mode collapse into the majority \texttt{Safe} class. For CRF-based models on SmolLM2 and in all cross-model settings, the emission logits are additionally scaled by the class-weight vector prior to Viterbi decoding to reinforce the rare onset state. This emission scaling was not applied in the TinyLlama and Mistral self-evaluation runs.

\paragraph{Environment}
Base generative models were run in \texttt{bfloat16}, or 8-bit quantization for Mistral-7B, prior to extracting \texttt{float32} internal representations. Token-level NAS were masked using an Activation Normalization Error threshold of 0.2.

\subsection{Cross-Model Supervision: Training Strategies}

For cross-model evaluation, external supervisors (Qwen-1.5B, Gemma-2-2B, Gemma-4) observe the generated text of SmolLM2 and Mistral-7B. The classifier architectures are unchanged, but two adjustments accommodate the cross-architectural domain gap and extreme sparsity.

\paragraph{Bridging the Latent Domain Gap}
Since the LayerNorm projector must translate a foreign representational space into BERT's native embeddings, we extend the BERT warm-up, keeping the encoder frozen for up to 5 epochs so the projector can learn the alignment before end-to-end fine-tuning at a reduced learning rate of $1\times10^{-5}$.

\paragraph{Mitigating Cross-Model Mode Collapse}
On the RAGTruth-Mistral data, the 0.48\% onset prevalence induced severe CRF mode collapse. We augment the objective with an auxiliary token-level Cross-Entropy loss on the raw emissions, penalizing missed onsets with \texttt{B-HALL} weights of 100--150, and scale this auxiliary term by a multiplier $\lambda \in [1, 3]$ so the linear projections fire on onsets independently of the global sequence probability.

\begin{table}[ht]
\centering
\small
\resizebox{\columnwidth}{!}{%
\begin{tabular}{@{}llp{4.2cm}@{}}
\toprule
\textbf{Cross-Model Hyperparameter} & \textbf{Value} & \textbf{Scope} \\ \midrule
Auxiliary \texttt{B-HALL} weight & 100 -- 150 & RAGTruth-Mistral onset penalty. \\
Auxiliary multiplier ($\lambda$) & 1 -- 3 & Token penalty vs.\ sequence logic. \\
BERT unfreeze epoch & 5 & Delayed for projector alignment. \\
Post-warmup BERT LR & $1\times10^{-5}$ & Stable end-to-end tuning. \\
Gradient accumulation & 4 steps & Memory boundary. \\ \bottomrule
\end{tabular}%
}
\caption{Training hyperparameters specific to the cross-model evaluation framework.}
\label{tab:cross_training_hyperparameters}
\end{table}

\paragraph{Data Separation Integrity}
All datasets were partitioned into 80/20 train/test splits prior to any hidden-state extraction or critical-layer analysis. The supervisor's critical layers ($\mathcal{C}_{critical}$) and NAS were computed exclusively on the training split, keeping the projector and classifier blind to the test distribution.

\subsection{Custom PsiloQA Generation and Verification}
To construct the custom SmolLM2 dataset, we used the base question prompts from PsiloQA \citep{rykov2025modelslielearnmultilingual}, inducing responses from SmolLM2 with a standard zero-shot QA template and no retrieval augmentation, so the model relied solely on its parametric memory. Generated responses were compared against the gold-standard answers, and spans that contradicted the reference or introduced fabricated entities were tagged under the strict BIO scheme.

\paragraph{Cross-Tokenizer Label Alignment}
Because our cross-model setting requires the same generated text to be labeled under different tokenizers, we anchor all annotations to character offsets rather than token identities. Hallucination spans are first marked at the character level against the gold reference, independent of any model's tokenization. For each model, we then retokenize the same text and recompute the BIO labels directly from that tokenizer's own offset mapping, assigning \texttt{B-HALL} to the first token overlapping a span and \texttt{I-HALL} to the remainder. Consequently, the generator and each observer receive their own internally consistent label sequence over their own tokens, and no cross-tokenizer projection is required. Partial overlaps are resolved by inclusion, so any token intersecting a hallucinated character range is labeled as part of that span.

\subsection{Development Environment}
All experiments were conducted on a 64-bit Linux workstation with an Intel Core i7-14700K (20 cores, 28 threads, 5.60 GHz) and 64 GB of RAM. Hidden-state extraction and model training used a single NVIDIA RTX 4000 Ada Generation GPU with 20 GB of GDDR6 VRAM, on NVIDIA driver 580.126.18 with CUDA 13.0, orchestrated within Jupyter.

\subsection{Artifact Licensing and Usage Terms}
The datasets and models used in this work are publicly available for academic research. PsiloQA and RAGTruth were used solely for non-commercial hallucination research, and the model families (SmolLM2, TinyLlama, Mistral, Qwen, Gemma) were accessed under their respective open or permissive licenses. No proprietary datasets were redistributed or modified, and any future release of code or processed artifacts will reference the original licenses and repositories.
\end{document}